\documentclass[a4paper]{article}

\usepackage{INTERSPEECH2021}
\usepackage{cite}
\usepackage{microtype}
\usepackage[super]{nth}

\newcommand{\argmax}{\operatornamewithlimits{argmax}}

\title{Reducing Exposure Bias in Training Recurrent Neural Network Transducers}
\name{Xiaodong Cui, Brian Kingsbury, George Saon, David Haws, Zoltan Tuske}
\address{IBM Research AI \\ IBM T. J. Watson Research Center, Yorktown Heights, New York, USA}
\email{\{cuix,bedk,gsaon,dhaws\}@us.ibm.com, zoltan.tuske@ibm.com}

\begin{document}

\maketitle
\begin{abstract}
When recurrent neural network transducers (RNNTs) are trained using the typical maximum likelihood criterion, the prediction network is trained only on ground truth label sequences. This leads to a mismatch during inference, known as exposure bias, when the model must deal with label sequences containing errors. In this paper we investigate approaches to reducing exposure bias in training to improve the generalization of RNNT models for automatic speech recognition (ASR). A label-preserving input perturbation to the prediction network is introduced. The input token sequences are perturbed using SwitchOut and scheduled sampling based on an additional token language model. Experiments conducted on the 300-hour Switchboard dataset demonstrate their effectiveness. By reducing the exposure bias, we show that we can further improve the accuracy of a high-performance RNNT ASR model and obtain state-of-the-art results on the 300-hour Switchboard dataset.
\end{abstract}
\noindent\textbf{Index Terms}: automatic speech recognition, end-to-end model,  RNN transducer, exposure bias, scheduled sampling

\section{Introduction}
\label{sec:intro}

End-to-end (E2E) automatic speech recognition (ASR) systems based on deep neural networks (DNNs) have made great progress in the past decade and have become more and more dominant in modern ASR. Compared to the conventional hybrid DNN ASR systems \cite{Hinton_DNNSPM,Dahl_CDDNN,Xiong_ASRParity,Saon_2017HumanParity} where output units represent context-dependent hidden Markov model (HMMs) states\cite{Bourlard_hybridasr}, E2E systems directly map an input acoustic sequence to an output text sequence. Therefore, pronunciation, acoustic and language modeling are carried out in the same framework. E2E models also usually require no hard frame level alignment, which greatly simplifies the ASR training pipeline. A broad variety of network architectures have been proposed for E2E systems in literature, notably connectionist temporal classification (CTC) \cite{Graves_CTC}, attention-based encoder-decoder (AED) \cite{Chan_LAS,Bahdanau_AED,Chorowski_AED}, recurrent neural network transducer (RNNT) \cite{Graves_RNNT,Graves_RNNASR,He_StreamingRNNT,Saon_RNNT} and self-attention-based transformer \cite{Vaswani_transformer}.

In recent years, RNNT models have emerged as a promising E2E ASR framework. They achieve competitive performance and in the meantime are streaming friendly. This makes them attractive in real-world deployments and hence they have received increasing attention in the community. In this paper, we are focused on improving RNNT ASR performance by reducing exposure bias during training. Exposure bias, a generic issue in text generation, arises due to the training-generation discrepancy. It happens when training is conducted with ground truth labels, while generation is conducted with errorful label sequences. In RNNT, the RNN prediction network predicts the next token conditioned on a history of previous tokens. In training, the prediction network is always fed with ground truth label sequences. However, errors may occur in decoding. Therefore the prediction of the next token is conditioned on a history contaminated with errors. This mismatch in training and decoding gives rise to exposure bias in RNNT models and hurts their generalization under test conditions.

Scheduled sampling \cite{Bengio_schep} is a representative technique for mitigating exposure bias. It has been used in E2E ASR frameworks such as AED models \cite{Chiu_S2S,Tuske_SWBSOTA} and self-attention-based transformers \cite{Zhou_ParaSchep}. Given the generative nature of the encoder-decoder architecture, scheduled sampling is relatively straightforward to realize in these frameworks. In RNNT, however, the prediction network essentially provides a ``soft'' token alignment of the acoustic feature sequence to compute the transition probabilities on the time-token lattice. Since it is not a generative architecture, it is not obvious how conventional scheduled sampling can be applied. Few efforts have been published on using scheduled sampling for training RNNT models. Scheduled sampling in \cite{Bengio_schep} can be viewed as a label-preserving perturbation of token history to introduce uncertainty during training. The perturbation is carried out such that the token history on which the next token prediction is conditioned is close to that observed in decoding so that the gap between the two is reduced. In the same light, we investigate in this paper label-preserving perturbation of input token sequences to the prediction network in RNNT to reduce the exposure bias. Specifically, we introduce perturbation based on SwitchOut \cite{Wang_Switchout} and perturbation based on scheduled sampling from a token language model (LM). Experiments are carried out on the 300-hour Switchboard (SWB300) dataset. We show that the perturbation of input token sequences is helpful in reducing exposure bias. It can improve over a high-performance RNNT ASR model with state-of-the-art word error rates (WERs).

The remainder of the paper is organized as follows. Section \ref{sec:rnnt} describes the RNNT framework. Section \ref{sec:pertb} introduces the two perturbation techniques to the input of the RNNT prediction network, namely, Switchout and scheduled sampling based on a token LM. Experimental results on SWB300 are reported in Section \ref{sec:exp} followed by a discussion in Section \ref{sec:dis}. Section \ref{sec:sum} concludes the paper with a summary.

\section{RNN Transducers}
\label{sec:rnnt}

Following the notation in \cite{Graves_RNNT}, let $\bm{x}=(x_{1},x_{2},\cdots,x_{T})$ be the input acoustic sequence of length $T$ where $x_{t} \in \mathcal{X}$ and $\bm{y}=(y_{1},y_{2},\cdots,y_{U})$ be the output token sequence of length $U$ where $y_{u} \in \mathcal{Y}$. $\mathcal{X}$ and $\mathcal{Y}$ are input and output spaces, respectively. Define the extended output space
\begin{align}
    \mathcal{\bar{Y}} = \mathcal{Y}\cup\varnothing
\end{align}
where $\varnothing$ represents a null output.

RNNT evaluates the conditional distribution of the output sequence given the input sequence
\begin{align}
    \text{Pr}(\bm{y}\in\mathcal{Y}^{*}|\bm{x}) = \sum_{\bm{a}\in\mathcal{B}^{-1}(\bm{y})}\text{Pr}(\bm{a}|\bm{x})  \label{eqn:loss}
\end{align}
where $\bm{a}\in\bar{\mathcal{Y}}^{*}$ are alignments of acoustic sequence $\bm{x}$ against the token sequence $\bm{y}$ with the null output symbol $\varnothing$ and $\mathcal{B}:\bar{\mathcal{Y}}^{*}\mapsto\mathcal{Y}^{*}$ is the function that removes null symbols from the alignments.

The acoustic features $x_{t}$ are embedded in a latent space by a bi-directional long short-term memory (LSTM) network \cite{Hochreiter_LSTM}, referred to as the transcription network $\mathcal{T}$
\begin{align}
      f_{t} & = \mathcal{T}(\bm{x}_{1:T},t).   \label{eqn:transnet}
\end{align}
The label tokens $y_{u}$ are embedded in a latent space by a uni-directional LSTM network, referred to as the prediction network $\mathcal{P}$:
\begin{align}
      g_{u} & = \mathcal{P}(\bm{y}_{[1:u-1]},u).  \label{eqn:prednet}
\end{align}

Given the acoustic embedding $f_{t}$ and the label token embedding $g_{u}$, the predictive output probability at $(t,u)$ is implemented as
\begin{align}
  p(\cdot|t,u) = \text{softmax}[\mathbf{W}^{\text{out}}\text{tanh}(\mathbf{W}^{\text{enc}}f_{t}\odot\mathbf{W}^{\text{pred}}g_{u}+b)]. \label{eqn:outp}
\end{align}
In Eq. \ref{eqn:outp}, matrices $\mathbf{W}^{\text{enc}}$ and $\mathbf{W}^{\text{pred}}$ are linear transforms that project the acoustic embedding $f_{t}$ and the label token embedding $g_{u}$ into the same joint latent space where element-wise multiplication is performed\footnote{Following the implementation in \cite{Saon_RNNT} due to observed superior performance on multiple datasets.} and a hyperbolic tangent ($tanh$) nonlinearity is applied. Finally, it is projected to the output space by a linear transform $\mathbf{W}^{\text{out}}$ and normalized by softmax, producing a predictive probability estimate.

The parameters of RNNT $\theta$ are optimized using the maximum likelihood criterion:
\begin{align}
     \theta^{*} = \argmax_{\theta} \log \text{Pr}(\bm{y}|\bm{x};\theta). \label{eqn:ml}
\end{align}
The conditional likelihood in Eq. \ref{eqn:loss} over all possible alignments can be evaluated using the forward-backward algorithm:
\begin{align}
    \text{Pr}(\bm{y}\in\mathcal{Y}^{*}|\bm{x}) = \sum_{(t,u):t+u=n}\alpha(t,u)\beta(t,u) \label{eqn:fwdbwd}
\end{align}
$\forall n: 1 \leq n \leq U+T$ where $\alpha(t,u)$ and $\beta(t,u)$ are forward and backward variables that can be computed recursively on the lattice.

 Optimization is typically based on stochastic gradient descent or its variants with back-propagation. Once the RNNT model is trained, decoding is carried out based on beam search \cite{Graves_RNNT,Saon_RNNTdecoding}.

\section{Input Perturbation of the Prediction Network}
\label{sec:pertb}

In standard RNNT training, the prediction network is always fed with the ground truth label token sequences. In Eq. \ref{eqn:prednet}, the LSTM $\mathcal{P}$ always uses the previous ground truth token history $\bm{y}_{[1:u-1]}$ to recurrently encode the current token $\bm{y}_{u}$. However, in decoding, the input sequence to the prediction network, $\hat{\bm{y}}=\{\hat{y}_{1}, \hat{y}_{2}, \cdots, \hat{y}_{U}\}$, is composed of token hypotheses. As a result, the prediction network has to encode the current token $\hat{y}_{u}$ using the history $\hat{\bm{y}}_{[1:u-1]}$ which may contain errors. This discrepancy between the training and decoding will incur the mismatch of the output probability in Eq. \ref{eqn:outp} and consequently affect the probability diffusion over the lattice.

In this section, we introduce label-preserving perturbation to the input label token sequences to the prediction network. Specifically, we introduce a perturbation $\sigma(\cdot)$
\begin{align}
     \sigma(\bm{y}) \rightarrow \tilde{\bm{y}}
\end{align}
to inject errors into the ground truth sequence. The resulting token sequence $\tilde{\bm{y}}$ perturbs the token embedding and also alters the alignment $\bm{a}$ in Eq. \ref{eqn:loss}
\begin{align}
    \text{Pr}(\bm{y}|\tilde{\bm{y}},\bm{x}) = \sum_{\bm{a}\in\mathcal{B}^{-1}(\tilde{\bm{y}})}\text{Pr}(\bm{a}|\bm{x}).  \label{eqn:ssloss}
\end{align}
In the meantime, since it is label preserving the loss in Eq. \ref{eqn:ml} is still optimized under the likelihood of ground truth $\bm{y}$.

In the sequel, we investigate two perturbation strategies, SwitchOut \cite{Wang_Switchout} and scheduled sampling \cite{Bengio_schep}.

\subsection{SwitchOut}
\label{sec:switchout}

Given an input token sequence $\bm{y}$, SwitchOut randomly corrupts a number of tokens in $\bm{y}$.

First, the number of tokens to be corrupted, $\hat{n}$, is sampled according to the following distribution
\begin{align}
  p(n) = \frac{e^{-\frac{n}{\tau}}}{\sum_{n'=0}^{|\bm{y}|}e^{-\frac{n'}{\tau}}}
\end{align}
where $\tau$ is a temperature parameter controlling the ``flatness'' of the distribution.

Define a Bernoulli random variable
\begin{align}
    \gamma \sim \text{Bernoulli}(\hat{n}/|\bm{y}|).
\end{align}
which is fixed for each token sequence $\bm{y}$.
For each token $y_{u}$ in $\bm{y}$, $u \in \{1,2, \cdots, U\}$,
\begin{align}
   \tilde{y}_{u} =
     \begin{cases}
       v\in{\cal Y}, v \neq y_{u}, &  \gamma = 1 \\
       y_{u}, &  \gamma = 0  \\
     \end{cases}
 \end{align}

\subsection{Scheduled Sampling}
\label{sec:schep}

SwitchOut introduces uncertainty to the ground truth sequence in a random fashion, which does not consider the history of tokens when making the perturbation. In decoding, every token is a predictive outcome of the previous history, and the prediction network implicitly learns the dependency similar to a language model. In order for the training to be closer to the decoding, we leverage a token LM to perturb the next token given the history.

First, an LSTM LM $\tilde{p}(z_{u}|\bm{z}_{[1:u-1]})$ is trained on the token sequences of the training set. Suppose $\bm{y}$ is the ground truth token sequence and $\tilde{\bm{y}}$ the perturbed. Define a Bernoulli random variable
\begin{align}
    \gamma \sim \text{Bernoulli}(p).
\end{align}
where $p$ is the teacher forcing probability.

To make the prediction of the next token in the perturbed sequence, it either sticks to the ground truth or samples from the token LM with probability $p$:
\begin{align}
  \tilde{y}_{u} =
    \begin{cases}
      y_{u}, &  \gamma = 1  \\
      z_{u} \sim \tilde{p}(z_{u}|\tilde{\bm{y}}_{[1:u-1]}), &  \gamma = 0
    \end{cases}
\end{align}
When sampling from $\tilde{p}(z_{u}|\bm{z}_{[1:u-1]})$, the predicted next token is uniformly selected from the top $k$ token candidates given the history $\tilde{\bm{y}}_{[1:u-1]}$, where $k$ is a hyper-parameter. When $k=1$, the most likely token given the history is always chosen.

\section{Experiments}
\label{sec:exp}

\subsection{Setup}

\noindent\textbf{Dataset} \ \ The experiments are conducted on the SWB300 dataset \cite{Godfrey_SWB,Cieri_Fisher}. Test sets include a 2.1-hour switchboard (SWB) set and a 1.6-hour call-home (CH) set from the NIST Hub5-2000 test set.  The data preparation pipeline follows the Kaldi \cite{Povey_Kaldi} s5c recipe.

\noindent\textbf{Architecture} \ \  There are 6 bi-directional LSTM layers in the transcription network with 1,280 cells in each layer (640 cells in each direction). The prediction network is a uni-directional LSTM consisting of a single layer with 1,024 cells. Both the acoustic embedding after the transcription network and the label embedding after the prediction network are projected down to a 256-dimensional latent space where they are combined by element-wise multiplication in the joint network. After a hyperbolic tangent nonlinearity followed by a linear transform, the topmost softmax layer has 46 output units which correspond to 45 characters and the null symbol.

\noindent\textbf{Acoustic features} \ \ The acoustic features are 40-dimensional logMel features after conversation side based mean and variance normalization plus their first and second order derivatives. The features are extracted every 10ms, but every two adjacent frames are concatenated, which amounts to a frame downsampling from 10ms to 20ms. This gives rise to the final 240-dimensional acoustic input to the transcription network.

\noindent\textbf{Regularization} \ \ The 300-hour training data is first augmented by speed and tempo perturbation \cite{Ko_SpeedTempo} to produce additional 4 replicas of the training data. On top of that, two additional data augmentation techniques are conducted on the fly when loading the training data. One is sequence noise injection \cite{Saon_mixup}, where a training utterance is artificially corrupted by adding a randomly selected downscaled training utterance from the training set, and the other is SpecAugment \cite{Park_SpecAug}, where the spectrum of a training utterance is randomly masked in blocks in both the time and frequency domains. The training is also regularized by dropout with a dropout rate of 0.25 for the LSTM and 0.05 for the embedding. In addition, DropConnect \cite{Wan_dropconnect} is applied with a rate of 0.25, which randomly zeros out elements of the LSTM hidden-to-hidden transition matrices. The transcription network is initialized by an LSTM model of the same configuration trained with CTC, following \cite{Audhkhasi_CTC}.

\noindent\textbf{Optimizer and Training Schedule} \ \  The RNNT is trained using \texttt{AdamW}~\cite{Loshchilov2019}, a version of the \texttt{Adam} optimizer~\cite{Kingma_adam} that adds decoupled weight decay. A long warmup and long hold training schedule is employed. The learning rate starts at 0.0002 in the first epoch and then linearly scales up to 0.002 in the first 10 epochs. It holds for another 6 epochs before being annealed by $\frac{1}{\sqrt{2}}$ every epoch after the \nth{17} epoch. The training finishes after 30 epochs.  In each epoch, the training utterances are provided in a sorted order, starting with short utterances. This amounts to a curriculum learning scheme that stabilizes the training early on with short utterances and then gradually increases the learning difficulty to longer utterances. The batch size is 250 which is uniformly distributed to 5 v100 GPUs (50 per each GPU).  Gradients are aggregated using \texttt{all-reduce}.

\noindent\textbf{Decoding} \ \ Inference uses alignment-length synchronous decoding~\cite{Saon_RNNTdecoding}, which only allows hypotheses with the same alignment length in the beam for the beam search. In all test cases, we measure the WERs with and without an external LM. When decoding with an external LM, the following density ratio LM fusion \cite{McDermott_LMdenratio} is used:
\begin{align}
   \bm{y}^{*}  =  \argmax_{\bm{y}\in\mathcal{Y}^{*}} & \left\{\log \text{Pr}(\bm{y}|\bm{x})-\mu \log \text{Pr}^{\text{src}}(\bm{y})  \right. \\ \nonumber
     & \left. \ \ \ \ + \lambda \log \text{Pr}^{\text{ext}}(\bm{y}) + \rho |\bm{y}| \right\}
\end{align}
where $\mu$ and $\lambda$ are the weights to balance the contribution of the source and external LMs, respectively, and $\rho$ for the label length reward $|\bm{y}|$. The external LM is trained on a target domain corpus (Fisher and Switchboard) and the source LM is trained only on the training transcripts.

Most of the above experimental settings and the training recipe follow those used in \cite{Saon_RNNT}.

\subsection{Results}

\noindent\textbf{SwitchOut}  \ \ Table \ref{tab:sw} shows the WERs of the RNNT baseline and RNNT after prediction network input perturbation using SwitchOut. The WERs for the RNNT baseline are 11.6\% on average for SWB and CH without an external LM and 10.2\% on average with the external LM using density ratio LM fusion. WERs for SwitchOut perturbation under various temperatures, $\tau$, are presented, among which $\tau=0.1$ gives the best performance. Under this condition, Switchout perturbation gives 0.2\% absolute improvement over the baseline in both test conditions.

\begin{table}[tbh]
\centering
\caption{WERs of baseline and SwitchOut perturbation under various temperatures $\tau$.}\label{tab:sw}
\begin{tabular}{ c | c c c | c c c} \hline
               &   \multicolumn{3}{c|}{w/o LM}   &  \multicolumn{3}{c}{w/ LM}   \\ \hline\hline
               &    swb    &    ch    &   avg   &   swb   &   ch   &   avg      \\ \hline
  baseline     &    \textbf{7.7}    &   \textbf{15.5}   &  \textbf{11.6}   &   \textbf{6.5}   &  \textbf{13.9}  &  \textbf{10.2}    \\ \hline
  $\tau=0.1$    &    \textbf{7.5}    &   \textbf{15.3}   &  \textbf{11.4}   &   \textbf{6.4}   &  \textbf{13.5}  &   \textbf{10.0}     \\ \hline
  $\tau=0.2$   &    7.3    &   15.5   &  11.4   &   6.2   &  13.8  &   10.0     \\ \hline
  $\tau=0.3$   &    7.5    &   15.5   &  11.5   &   6.3   &  13.8  &   10.1     \\ \hline
  $\tau=0.5$   &    7.5    &   15.6   &  11.6   &   6.4   &  13.9  &   10.2     \\ \hline
\end{tabular}
\end{table}

\noindent\textbf{Scheduled Sampling} \ \  Table \ref{tab:schesp} shows the WERs of the RNNT baseline, which is the same as Table \ref{tab:sw}, and RNNT under scheduled sampling with various configurations. The token LM is a single layer LSTM trained on the character sequences of the training transcripts.    There are two hyper-parameters under investigation in the experiments. One is the number of candidates (top$k$) from which the next token is sampled from the token LM.  The other hyper-parameter is the teacher forcing probability $p$. Experiments are carried out by varying the top$k$ and teacher forcing probability. It can be observed from the table that scheduled sampling is helpful to improve the recognition performance. The best result is given by $\text{top}k\!=\!3$ and $p\!=\!0.9$ where the average WER is 11.2\% without the external LM and $9.7\%$ with the external LM. This amounts to $0.4\%$ and $0.5\%$ absolute improvement under the two test conditions, respectively. It also outperforms the best result under SwitchOut in Table \ref{tab:sw} by 0.2\% absolute without using the external LM and 0.3\% absolute when using it. However, it is worth noting that scheduled sampling is more computationally demanding, as the sampling depends on the token history.

\begin{table}[tbh]
\centering
\caption{WERs of baseline and scheduled sampling under various top $k$ candidates and teacher forcing probability $p$.}\label{tab:schesp}
\begin{tabular}{ c | c c c | c c c} \hline
                       &   \multicolumn{3}{c|}{w/o LM}  &  \multicolumn{3}{c}{w/ LM}   \\ \hline\hline
                       &    swb    &    ch    &   avg   &   swb   &   ch   &   avg     \\ \hline
baseline    &    \textbf{7.7}    &   \textbf{15.5}   &  \textbf{11.6}   &   \textbf{6.5}   &  \textbf{13.9}  &  \textbf{10.2}    \\ \hline
  $\text{top}k\!=\!1$, $p\!=\!0.9$  &    7.3    &   15.4   &  11.4   &   6.3   &  13.7  &   10.0    \\ \hline
  $\text{top}k\!=\!2$, $p\!=\!0.9$  &    7.4    &   15.2   &  11.3   &   6.4   &  13.7  &   10.1    \\ \hline
  $\text{top}\!k=\!3$, $p\!=\!0.9$  &   \textbf{7.2}    &  \textbf{15.2}   &  \textbf{11.2}   &  \textbf{6.1}   &  \textbf{13.3}  &  \textbf{9.7}     \\ \hline
  $\text{top}k\!=\!5$, $p\!=\!0.9$  &    7.3    &   15.2   &  11.3   &   6.2   &  13.5  &   9.9     \\ \hline
  $\text{top}k\!=\!1$, $p\!=\!0.8$  &    7.5    &   15.3   &  11.4   &   6.3   &  13.6  &   10.0    \\ \hline
  $\text{top}k\!=\!2$, $p\!=\!0.8$  &    7.4    &   15.3   &  11.4   &   6.3   &  13.5  &   9.9      \\ \hline
\end{tabular}
\end{table}

\noindent\textbf{Input with i-vector} \ \ We also train RNNT models using input with i-vectors to further improve the baseline. The input features are a concatenation of the previous 240-dimensional features and speaker-dependent 100-dimensional i-vectors \cite{Dehak_ivec,Saon_IVECT}. Therefore, the input dimensionality in this case is 340. We reduce the hidden cells of the LSTM in the prediction network from 1,024 to 768. Table \ref{tab:ivec} presents the WERs of the baseline RNNT with i-vectors and the WERs under scheduled sampling with various top$k$ and teacher forcing probabilities. The WERs of the RNNT baseline with i-vector are 11.3\% on average without using the external LM and 9.7\% with the external LM. This matches the best performance reported in \cite{Saon_RNNT}, which is a state-of-art result on SWB300. As can be seen from the table, with the help of scheduled sampling, the best result is obtained when $\text{top}k\!=\!4$ and $p\!=\!0.8$: 11.0\% on average without the external LM and 9.6\% with the external LM. We also investigate an annealing strategy where scheduled sampling is used up to 25 epochs, after which it is lifted and only ground truth label sequences are used. The learning rate for the \nth{26} epoch is set to 1.2e-4 and annealed by $\frac{1}{\sqrt{2}}$ for each epoch afterwards. This can further improve the WERs when no external LM is used ($11.0\%\!\rightarrow\!10.9\%$), but with the external LM used the WERs stay the same (9.6\%).

\begin{table}[tbh]
\centering
\caption{WERs of baseline and scheduled sampling under various top $k$ candidates and teacher forcing probability. Input acoustic features include i-vector.}\label{tab:ivec}
\begin{tabular}{ c | c c c | c c c} \hline
                       &   \multicolumn{3}{c|}{w/o LM}  &  \multicolumn{3}{c}{w/ LM}   \\ \hline\hline
                       &    swb    &    ch    &   avg   &   swb   &   ch   &   avg     \\ \hline
   baseline            &    \textbf{7.3}    &   \textbf{15.3}   &  \textbf{11.3}   &   \textbf{6.0}   &  \textbf{13.4}  &   \textbf{9.7}     \\ \hline
  $\text{top}k\!=\!1$, $p\!=\!0.9$  &    7.3    &   15.2   &  11.3   &   6.2   &  13.4  &   9.8     \\ \hline
  $\text{top}k\!=\!3$, $p\!=\!0.9$  &    7.3    &   15.1   &  11.2   &   6.3   &  13.4  &   9.9     \\ \hline
  $\text{top}k\!=\!5$, $p\!=\!0.9$  &    7.1    &   15.1   &  11.1   &   6.1   &  13.3  &   9.7     \\ \hline
  $\text{top}k\!=\!1$, $p\!=\!0.8$  &    7.4    &   15.0   &  11.2   &   6.1   &  13.5  &   9.8     \\ \hline
  $\text{top}k\!=\!2$, $p\!=\!0.8$  &    7.4    &   15.0   &  11.2   &   6.2   &  13.4  &   9.8     \\ \hline
  $\text{top}k\!=\!3$, $p\!=\!0.8$  &    7.1    &   15.0   &  11.1   &   6.0   &  13.2  &   9.6   \\
  \ \ + anneal         &    7.0    &   14.7   &  10.9   &   6.0   &  13.2  &   9.6     \\ \hline
  $\text{top}k\!=\!4$, $p\!=\!0.8$ &   \textbf{7.0}    &  \textbf{15.0}   &  \textbf{11.0}   &   \textbf{6.0}   &  \textbf{13.2}  &  \textbf{9.6}   \\
  \ \ + anneal          &    \textbf{6.9}    &  \textbf{14.9}   &  \textbf{10.9}   &   \textbf{5.9}   &  \textbf{13.2}  &   \textbf{9.6}     \\ \hline
  $\text{top}k\!=\!5$, $p\!=\!0.8$  &    6.9    &   15.4   &  11.2   &   6.0   &  13.4  &   9.7   \\   \hline
 \end{tabular}
\end{table}

\section{Discussion}
\label{sec:dis}

Both SwitchOut and scheduled sampling are shown to improve the performance of RNNT models. While SwitchOut perturbs the label sequence by uniformly sampling tokens from ${\cal Y}$ to replace the original token, scheduled sampling samples the next predicted token using a token LM given the observed history. Hence, scheduled sampling outperforms SwitchOut in this case to reduce exposure bias. This superior performance, however, comes with a higher computational cost in the training as one has to recurrently generate the history with perturbed tokens for each label sequence. Once the model is trained, the inference time of both strategies is no different from the RNNT baseline.

The RNNT baseline with i-vector in Table \ref{tab:ivec} with 6.0/13.4/9.7 WERs is one of the best single ASR systems with state-of-the-art performance \cite{Saon_RNNT}. With scheduled sampling, the WERs can be further improved to 5.9/13.2/9.6, which gives one of the best results on the SWB300 dataset. If we reduce the batch size from 250 to 64, a slightly better performance is obtained (6.0/13.0/9.5). A smaller batch size may give rise to noisier stochastic gradients, which is helpful to search for a better local optimum. But the training time is significantly longer. Table \ref{tab:wer} summarizes the performance of best single models reported in the literature. 

\begin{table}[tbh]
\centering
\caption{WERs of single models reported in literature on SWB300.}\label{tab:wer}
\begin{tabular}{ l | c | c c c} \hline
     system                                  &  model    &   swb   &   ch   &   avg   \\ \hline\hline
    Park et al. / 2019 \cite{Park_SpecAug}   &  AED      &   6.8   &  14.1  &   10.5  \\ \hline
    Irie et al. / 2019 \cite{Irie_CrossSent} & Hybrid    &   6.7   &  12.9  &   9.8   \\ \hline
    Tuske et al. / 2020 \cite{Tuske_SWBSOTA} &  AED      &   6.4   &  12.5  &   9.5   \\ \hline
    Saon  et al. / 2021 \cite{Saon_RNNT}     &  RNNT     &   6.3   &  13.1  &   9.7   \\ \hline
    This work (batch size 250)               &  RNNT     &   5.9   &  13.2  &   9.6   \\ \hline
    This work (batch size 64)                &  RNNT     &   6.0   &  13.0  &   9.5   \\ \hline
\end{tabular}
\end{table}

By comparing Table \ref{tab:schesp} and Table \ref{tab:ivec}, one can observe that an RNNT trained with scheduled sampling without using i-vector in the input (9.7\%) matches the performance of an RNNT baseline using i-vector in the input (9.7\%). Scheduled sampling can further improve the latter (9.6\%). This observation is helpful, as computing i-vectors is computationally expensive and needs sufficient data from a single speaker. Therefore, in some real-world applications, including i-vectors in input acoustic features is either computationally inefficient or technically infeasible, especially when ASR is carried out in a streaming mode.

\section{Summary}
\label{sec:sum}

In this paper we introduced label-preserving perturbation to the input to the prediction network of an RNNT to reduce exposure bias. Two perturbation strategies were investigated, SwitchOut and scheduled sampling. SwitchOut randomly corrupts the ground truth label token sequence, while scheduled sampling draws the next token based on an additional token LM given the history. Both strategies have been shown to be helpful in reducing the WERs of a high-performance RNNT ASR model. Input perturbation based on scheduled sampling obtains one of best results on the SWB300 dataset.

\bibliographystyle{IEEEtran}

\bibliography{rnnt_schesp}

\end{document}